\documentclass[final,5p,times,hidelinks,twocolumn]{elsarticle}

\usepackage[hidelinks]{hyperref}

\usepackage[english]{babel}			
\usepackage{blindtext} 				
\usepackage{color,soul}				
\usepackage{amsmath}				
\usepackage{diagbox}				
\usepackage{subcaption}				
\usepackage{multirow}				
\usepackage[table,xcdraw]{xcolor}	

\makeatletter
\def\ps@pprintTitle{%
 \let\@oddhead\@empty
 \let\@evenhead\@empty
 \def\@oddfoot{}%
 \let\@evenfoot\@oddfoot}
\makeatother

\biboptions{authoryear}

\begin{document}

\begin{frontmatter}

\title{Automated sub-cortical brain structure segmentation combining spatial and deep convolutional features}

\author[]{Kaisar Kushibar\corref{mycorrespondingauthor}}
\ead{kaisar.kushibar@udg.edu}
\author[]{Sergi Valverde\corref{mycorrespondingauthor}}
\cortext[mycorrespondingauthor]{These authors contributed equally to this work}
\ead{sergio.valverde@udg.edu}
\author[]{Sandra Gonz\'alez-Vill\`a}
\ead{sgonzalez@eia.udg.edu}
\author[]{Jose Bernal}
\ead{jose.bernal@udg.edu}
\author[]{Mariano Cabezas}
\ead{mariano.cabezas@udg.edu}
\author[]{Arnau Oliver}
\ead{aoliver@eia.udg.edu}
\author[]{Xavier Llad\'o}
\ead{xavier.llado@udg.edu}

\address{Institute of Computer Vision and Robotics, University of Girona.\\
Ed. P-IV, Campus Montilivi, University of Girona, 17003 Girona (Spain).}



\begin{abstract}
Sub-cortical brain structure segmentation in Magnetic Resonance Images (MRI) has attracted the interest of the research community for a long time because morphological changes in these structures are related to different neurodegenerative disorders. However, manual segmentation of these structures can be tedious and prone to variability, highlighting the need for robust automated segmentation methods. In this paper, we present a novel convolutional neural network based approach for accurate segmentation of the sub-cortical brain structures that combines both convolutional and prior spatial features for improving the segmentation accuracy. In order to increase the accuracy of the automated segmentation, we propose to train the network using a restricted sample selection to force the network to learn the most difficult parts of the structures. We evaluate the accuracy of the proposed method on the public MICCAI 2012 challenge and IBSR 18 datasets, comparing it with different available state-of-the-art methods and other recently proposed deep learning approaches. On the MICCAI 2012 dataset, our method shows an excellent performance comparable to the best challenge participant strategy, while performing significantly better than state-of-the-art techniques such as FreeSurfer and FIRST. On the IBSR 18 dataset, our method also exhibits a significant increase in the performance with respect to not only FreeSurfer and FIRST, but also comparable or better results than other recent deep learning approaches. Moreover, our experiments show that both the addition of the spatial priors and the restricted sampling strategy have a significant effect on the accuracy of the proposed method. In order to encourage the reproducibility and the use of the proposed method, a public version of our approach is available to download for the neuroimaging community.  

\end{abstract}

\begin{keyword}
Brain, MRI\sep sub-cortical structures\sep segmentation\sep convolutional neural networks
\end{keyword}

\end{frontmatter}

\section{Introduction}
\label{sec:introduction}
Brain structure segmentation in Magnetic Resonance Images (MRI) is one of the major interests in medical practice due to its various applications, including pre-operative evaluation and surgical planning, radiotherapy treatment planning, longitudinal monitoring for disease progression or remission \citep{kikinis1996digital,phillips2015prospective,pitiot2004expert}. The sub-cortical structures (i.e. thalamus, caudate, putamen, pallidum, hippocampus, amygdala, and accumbens) have attracted the interest of the research community for a long time, since their morphological changes are frequently associated with psychiatric and neurodegenerative disorders and could be used as biomarkers of some diseases \citep{debernard2015deep,mak2014subcortical}. Therefore, segmentation of sub-cortical brain structures in MRI for quantitative analysis has a major clinical application. However, manual segmentation of MRI is extremely time consuming and hardly reproducible due to inter- and intra- variability of the operators, highlighting the need for automated accurate segmentation methods.


Recently, \cite{gonzalez2016review}, reviewed different approaches for brain structure segmentation in MRI.
One of the commonly used automatic brain structure segmentation tools in medical practice is FreeSurfer\footnote{\url{https://surfer.nmr.mgh.harvard.edu/}}, which uses non-linear registration and an atlas-based segmentation approach \citep{fischl2002whole}. Another classical approach, also popular in the medical community, is the method proposed by \cite{patenaude2011first} -- FIRST, which is included into the publicly available software FSL\footnote{\url{https://fsl.fmrib.ox.ac.uk/fsl/fslwiki}}. This method uses the principles of Active Shape \citep{cootes1995active_shape_models} and Active Appearance Models \citep{cootes2001active_appearance_model} that are put within a Bayesian framework, allowing to use the probabilistic relationship between shape and intensity to its full extent.

In recent years, deep learning methods, in particular, Convolutional Neural Networks (CNN), have demonstrated a state-of-the-art performance in many computer vision tasks such as visual object detection, classification and segmentation \citep{krizhevsky2012imagenet,he2016microsoft,szegedy2015googlenet,girshick2014rich}. Unlike handcrafted features, CNN methods learn from observed data \citep{lecun1998gradient} making the features more relevant to a task. Therefore, CNNs are also becoming a popular technique applied in medical image analysis. There have been many advances in the application of deep learning in medical imaging such as expert-level performance in skin cancer classification \citep{esteva2017dermatologist}, high rate detecting cancer metastases \citep{liu2017detecting}, Alzheimer's disease classification \citep{sarraf2016deepad}, and spotting early signs of autism \citep{hazlett2017early}.


Some CNN methods have also been proposed for brain structure segmentation. One of the common techniques used in the literature is patch-based segmentation, where patches of a certain size are extracted around each voxel and classified using a CNN. Application of 2D, 3D, 2.5D patches (three patches from the orthogonal views of an MRI volume) and their combinations including multi-scale patches can be found in the literature for brain structure segmentation \citep{montana2015deep,bao2016multi,milletari2017hough,mehta2017brainsegnet}. Combining patches of different dimensions is done in a multi-path manner, where CNNs consist of different branches corresponding to each patch type. In contrast to patch-based CNNs, fully convolutional neural networks (FCNN) produce segmentation for a neighborhood of an input patch \citep{long2015fully}. \cite{shakeri2016sub} adapted the work of \cite{chen2014semantic} for semantic segmentation of natural images using FCNN. Moreover, 3D FCNNs, which segment a 3D neighborhood of an input patch at once, have been investigated by \cite{dolz20173d} and \cite{wachinger2017deepnat}. Although FCNNs show improvement in segmentation speed due to parallel segmentation of several voxels, it suffers from a high number of parameters in the network in comparison with patch-based CNNs. Also, it is common to apply post-processing methods to refine the final segmentation output. Inference of CNN-priors and statistical models such as Markov Random Fields and Conditional Random Fields \citep{lafferty2001crf} were used in the experiments of \cite{montana2015deep}, \cite{shakeri2016sub}, and \cite{wachinger2017deepnat}. A modified Random Walker based segmentation refinement has been also proposed by \cite{bao2016multi}. Apart from implicit information that is provided by the extracted patches from MRI volumes, explicit characteristics distinguishing spatial consistency have been studied. \cite{montana2015deep} included distances to centroids to their networks. \cite{wachinger2017deepnat} used the Euclidean and spectral coordinates computed from eigenfunctions of a Laplace-Beltrami operator of a solid 3D brain mask, to provide a distinctive perception of spatial location for every voxel. These kinds of features provide additional spatial information, however, extracting these explicit features from an unannotated MRI volume requires some preliminary operations to be attended (e.g. repetitive training of the network to compute initial segmentation mask).

From the reviewed literature, we have observed that most of the current deep learning approaches for sub-cortical brain structure segmentation focus on segmenting only the large sub-cortical structures (thalamus, caudate, putamen, pallidum). However, other important small structures (i.e. hippocampus, amygdala, accumbens), which are used for examining neurological disorders such as schizophrenia \citep{altshuler1998amygdala,lawrie2003structural}, anxiety disorder \citep{milham2005selective}, bipolar disorder \citep{altshuler1998amygdala}, Alzheimer \citep{fox1996presymptomatic}, etc., are not considered. In this work, we are presenting our CNN approach for segmenting all the sub-cortical structures. A recent approach of \cite{ghafoorian2017location} has been taken as a seminal work in our research. In their work, spatial features, provided by tissue atlas probabilities, were combined with 2D CNN features for segmenting White Matter Hyperintensities in MRI. In this paper, we are presenting a different 2.5D CNN architecture, i.e the three orthogonal views of the 3D volume, for segmenting the sub-cortical brain structures that combines spatial features in a similar way to \cite{ghafoorian2017location}. To the best of our knowledge, this is the first deep learning method incorporating atlas probabilities for sub-cortical brain structure segmentation.
Moreover, we propose a new sample selection technique to allow the neural network to learn to segment the most difficult areas of the structures in the images. We test the proposed strategy in two well-known datasets: MICCAI 2012\footnote{\url{https://masi.vuse.vanderbilt.edu/workshop2012}} \citep{landman2012miccai} and IBSR 18\footnote{\url{https://www.nitrc.org/projects/ibsr}}; and compare our results with the classical and recent CNN strategies for brain structure segmentation. Moreover, we make our method publicly available for the community, accessible online at \url{https://github.com/NIC-VICOROB/sub-cortical_segmentation}.

\section{Method}
\label{sec:methods}
\subsection{Input features}
in our method, we employ 2.5D patches to incorporate information from three orthogonal views of a 3D volume. In our case, each patch has a size of $32\times 32$ pixels. 3D patches provide more information of surroundings for the voxel that is being classified, but it is computationally and memory expensive. Thus, by using 2.5D patches, we approximate the information that is provided by a 3D patch in computational time and memory efficient manner.

Along with the appearance based features provided by the T1-w MRI, we employ spatial features extracted from a structural probabilistic atlas. In our experiments, we used the well-known Harvard-Oxford \citep{caviness1996harvard_oxford_atlas} atlas template in MNI152 space distributed with the FSL package \footnote{\url{https://fsl.fmrib.ox.ac.uk/fsl/fslwiki}}, which has been built using 47 young adult healthy brains. In our method, first, T1-w image of the MNI152 template is affine registered to T1-w image of the considered datasets using a block matching approach \citep{ourselin2000block}. Then, non-linear registration of the atlas template to subject volume is applied using fast free-form deformation method \citep{modat2010fast}. The deformation field obtained after the registration is used to move the probabilistic atlas into the subject space. Registration processes have been carried out using the well known and publicly available tool NiftyReg\footnote{\url{http://cmictig.cs.ucl.ac.uk/wiki/index.php/NiftyReg}}. Afterwards, vectors of size 15, corresponding to seven anatomical structures with left and right parts separately and background, were extracted from probabilistic atlas for every voxel and used as an input feature to train the network.

\subsection{CNN architecture}
\begin{figure}[tb]
\centering
\includegraphics[width=0.5\textwidth]{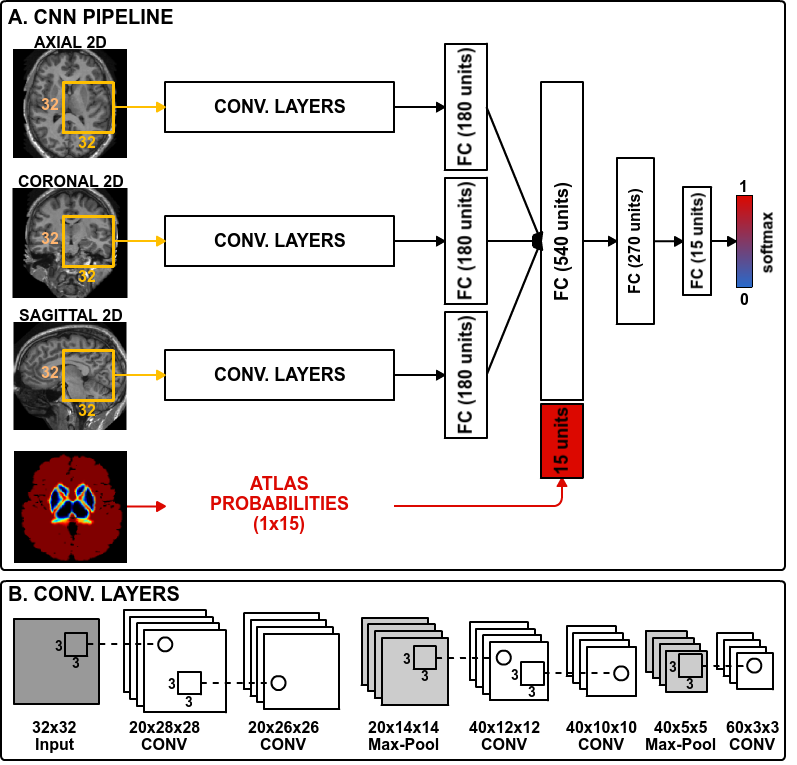}
\caption{The proposed 2.5D CNN architecture has three convolutional branches and a branch for spatial prior. 2D patches of size $32\times 32$ pixels are extracted from three orthogonal views of a 3D volume. Spatial prior branch accepts a vector of size 15 with atlas probabilities for each of the 14 structures and background.}
\label{fig:architecture_proposal}
\end{figure}

Figure~\ref{fig:architecture_proposal} illustrates our proposed CNN architecture. It consists of three branches corresponding to the patches extracted from axial, coronal, and sagittal views of a 3D volume, and one branch corresponding to the spatial priors. The branch for the spatial prior accepts a vector of size 15 with atlas probabilities for each structure and the background. The first three branches have the same organization of convolutional and max-pooling layers as shown in Figure~\ref{fig:architecture_proposal}(B). All the feature maps of the convolutional layers are passed through the Rectified Linear Unit (ReLU) activation function \citep{glorot2011deep}. For all the convolutional layers, kernels of size $3\times 3$ are set to make the CNN deep without losing in performance and bursting the number of parameters as it has been studied in \cite{simonyan2014very_deep}. Then, the outputs of the convolutional layers are flattened and followed by fully connected (FC) layers with 180 units each. Next, FC layers of each branch including atlas probabilities are fully connected to two consecutive FC layers with 540 and 270 units. The final classification layer has 15 units with the softmax activation function.

\subsection{CNN training}
For training our network, we extract 2.5D patches from the training set and using the provided ground truth labels we optimize the kernel and fully connected layer unit weights based on the loss function. In the proposed network we employ the categorical cross-entropy loss function, which is minimized using the Adam \citep{kingma2014adam} optimization method. This technique automatically controls the learning rate and uses moving averages of the parameters, which allows the step size to be effectively large and converge to optimal step size without tuning it manually.

\begin{figure*}[tb]
\centering
\begin{subfigure}[b]{0.20\textwidth}
  \includegraphics[width=\textwidth]{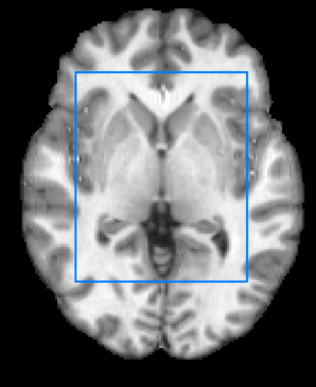}
  \caption{}
  \label{fig:sample_selection_t1_brain}
\end{subfigure}
\begin{subfigure}[b]{0.20\textwidth}
  \includegraphics[width=\textwidth]{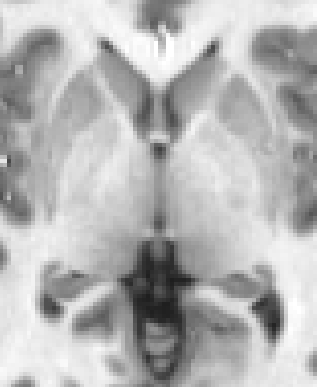}
  \caption{}
  \label{fig:sample_selection_t1}
\end{subfigure}
\begin{subfigure}[b]{0.20\textwidth}
  \includegraphics[width=\textwidth]{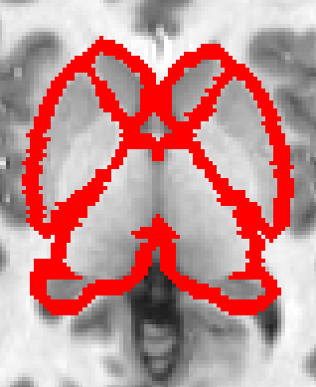}
  \caption{}
  \label{fig:sample_selection_bdr}
\end{subfigure}
\begin{subfigure}[b]{0.20\textwidth}
  \includegraphics[width=\textwidth]{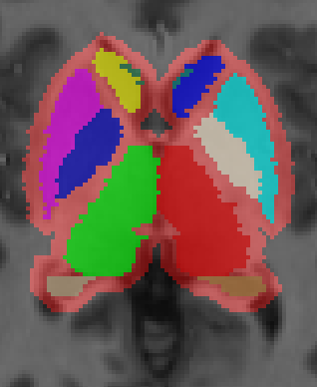}
  \caption{}
  \label{fig:sample_selection_gt}
\end{subfigure}
\caption{Negative sample selection from the boundaries of the target structures. (a) T1-w image with a rectangle representing the ROI; (b) T1-w ROI; (c) structure boundaries; (d) groundtruth labels with boundaries.}
\label{fig:negative_from_boundaries}
\end{figure*}

When training the CNN, it is important to take into account how the training samples are extracted from an image. Random selection of certain number of samples from an image is one of the common techniques in the literature. However, when it comes to the segmentation of the sub-cortical structures, the background (negative) samples turn out to be dispersed in the subject volume. Hence, it would lead to imperfect segmentation results on the borders of the structures, which are the most delicate areas to process due to the low contrast between the structure and the background. Therefore, we propose to extract the negative samples only from the structure boundaries as shown in Figure~\ref{fig:negative_from_boundaries}. In doing so, we force the network to learn only from the structure boundaries and dismiss other parts of the background.

The training sample selection is performed as follows: from all the available training images, we first select the positive samples from all the voxels from the 14 sub-cortical structures. Then, the same number of negative samples are randomly selected from the structure boundaries within five voxel distance, forming a balanced dataset of sub-cortical and boundary voxels. More details about batch size and number of epochs of the training process for the selected datasets will be given in Section~\ref{sec:results}.

\subsection{CNN testing}
To perform the segmentation of a new image volume, we extract all the patches from the image and predict class label probabilities using the trained CNN. Then, we assign a label corresponding to the maximum a posteriori probability for every input patch. Notice that knowing the order of the patch extraction is important to be able to reconstruct the final segmentation output. We also take advantage of the location of the sub-cortical structures, which are located in the central part of the brain. Due to the knowledge provided by the atlases, regions of interest (ROI) are automatically defined for all the subject volumes to achieve faster training and testing speeds.

Since the network has been trained with the negative samples extracted only from the structure boundaries, it will produce spurious outputs in unseen areas of the background when segmenting a testing volume. In order to overcome this issue, we apply a post-processing step, where for each class only the region with the biggest volume within the ROI is preserved.

\subsection{Implementation and technical details}
The proposed method has been implemented in the Python language\footnote{\url{https://www.python.org/}}, using Lasagne\footnote{\url{http://lasagne.readthedocs.io}} and Theano\footnote{\url{http://deeplearning.net/software/theano/}} \cite{bergstra2011theano} libraries. All experiments have been run on a GNU/Linux machine box running Ubuntu 16.04, with 32 GB RAM memory. CNN training has been carried out on a single TITAN-X GPU (NVIDIA corp, United States) with 12 GB RAM memory. The proposed method is currently available for downloading at our research website\footnote{\url{https://github.com/nic-vicorob/cnn_subcortical_segmentation}}.

\section{Results} \label{sec:results}
This section presents the results obtained by the proposed method on two datasets. The first dataset is the one provided in the MICCAI Multi-Atlas Labeling challenge\footnote{\url{https://masi.vuse.vanderbilt.edu/workshop2012}} \citep{landman2012miccai} and the second is a publicly available dataset from the Internet Brain Segmentation Repository\footnote{\url{https://www.nitrc.org/projects/ibsr}} (IBSR). Details of these datasets and the corresponding results will be given in Section~\ref{sec:miccai_results} and in Section~\ref{sec:ibsr_results}.

\subsection{Evaluation measures}
For evaluating the proposed method, we selected two metrics that are commonly used in the literature. These are overlap and spatial distance-based metrics, which show similarity and discrepancy of automatic and manual segmentations. The first measurement is Dice Similarity Coefficient (DSC) \citep{dice1945measures} defined as the following for automatic segmentation $A$ and manual segmentation $B$:
\begin{equation} \label{eq:dice}
	DSC(A,B) = \frac{2|A \cap B|}{|A|+|B|}.
\end{equation}
DSC measures the overlap of the segmentation with the ground truth on a scale between 0 and 1, where the former shows no overlap and the latter represents 100\% overlap with the ground truth.

For the spatial distance based metric, Hausdorff Distance (HD) is used in our experiments. This metric is defined as a function of the Euclidean distances between the voxels of $A$ and $B$ as:.
\begin{equation} \label{eq:hausdorff}
  \begin{gathered}
    HD(A, B) = \max(h(A, B), h(B, A)),\\
    h(A, B) = \max_{a\in A}\min_{b\in B}||a-b||.
  \end{gathered}
\end{equation}
In other words, HD is the maximum distance from all the minimum distances between boundaries of segmentation and boundaries of the ground truth.

Similarly to \cite{wachinger2017deepnat}, we used Wilcoxon signed-rank test to test the statistical significance of: 1) the differences in DSC and HD between our and state-of-the-art methods; and 2) the effect of using spatial features and the proposed sample selection technique.

\subsection{MICCAI 2012 Dataset} \label{sec:miccai_results}
This dataset consists of 35 T1-w MRI volumes split into 15 cases for training and 20 cases for testing. Manually segmented ground truth for each image is available as well, which contains 134 structures overall. In our experiments, we extracted 14 classes corresponding to seven sub-cortical structures with left and right parts separately. All the subject volumes have even voxel spacing of 1 mm$^3$ with a size of $256\times 256\times 256$ voxels in axial, sagittal, and coronal views respectively.

\subsubsection{Experimental details} \label{sec:experimental_details_miccai}
Skull-stripping was applied to extract the brain and cut out other parts appearing in the MRI such as eyes, skull, skin, and fat using the BET algorithm \citep{smith2002bet}. Then, the spatial intensity variations on the MRI volumes were corrected using a bias field correction algorithm -- N4ITK \citep{tustison2010n4itk}, which is included in the publicly available ITK\footnote{\url{https://itk.org/}} toolkit. Both preprocessing methods were run with default parameters.


In our experiments, we trained a single model using the available training set of 15 images, while we tested the other 20 images as provided in the original MICCAI 2012 Challenge. From the training set, we extracted around $1.5M$ ($750K$ of sub-cortical voxels and $750K$ of boundary voxels) sample patches of size $32\times 32$ pixels from three orthogonal views, where around $1.1M$ ($75\%$) were used for training and $400K$ samples for validation ($25\%$). The extracted patches were passed to the network for training in batches of size $128$. The network was trained for 200 epochs, while in order to prevent the network from over-fitting, we applied early stopping of the training process. The training process was automatically terminated when the validation accuracy did not increase after 20 epochs. 

\subsubsection{Comparison with other available methods}
The performance of the proposed approach is compared with widely used tools in medical practice -- FreeSurfer and FIRST. We also compared the performance of our method with the one of PICSL \citep{wang2013picsl} method, which is a multi-atlas based segmentation strategy that uses joint fusion technique with corrective learning. PICSL has been the winner of the MICCAI 2012 Challenge for brain structure segmentation and still shows the best results on this dataset. For the methods of FreeSurfer and FIRST, we used their default parameters to produce segmentation masks for the testing volumes. Accordingly, the training and testing split matches the configuration we used for evaluating the proposed method. We have to note that, with this dataset, there were no individually reported numerical results for each of the sub-cortical structure in other CNN based approaches.

\subsubsection{Results}
\begin{table*}[tb]
\scriptsize
\centering
\caption{MICCAI 2012 dataset results. Mean DSC $\pm$ standard deviation and HD $\pm$ standard deviation values for each structure obtained using FreeSurfer, FIRST, PICSL, and our method. Structure acronyms are: left thalamus (Tha.L), right thalamus (Tha.R), left caudate (Cau.L), right caudate (Cau.R), left putamen (Put.L), right putamen (Put.R), left pallidum (Pal.L), right pallidum (Pal.R), left hippocampus (Hip.L), right hippocampus (Hip.R), left amygdala (Amy.L), right amygdala (Amy.R), left accumbens (Acc.L), right accumbens (Acc.R) and average value (Avg.). Highest DSC and HD values for each structure are shown in bold.}
\label{table:mean_dsc_hd_miccai}
\begin{tabular}{|c|c|c|c|c|c|c|c|c|}
\hline
Method & \multicolumn{2}{c|}{FreeSurfer \cite{fischl2012freesurfer}} & \multicolumn{2}{c|}{FIRST \cite{patenaude2011first}} & \multicolumn{2}{c|}{PICSL \cite{wang2013picsl}} & \multicolumn{2}{c|}{Our method} \\ \hline
Structure & DSC & HD & DSC & HD & DSC & HD & DSC & HD \\ \hline \hline
Tha.L & 0.830$\pm$0.018 & 4.94$\pm$1.01 & 0.889$\pm$0.018 & 4.65$\pm$0.90 & 0.920$\pm$0.013 & \textbf{3.22$\pm$0.99} &\textbf{ 0.921$\pm$0.018} & 3.39$\pm$1.13 \\ \hline
Tha.R & 0.849$\pm$0.021 & 4.76$\pm$0.75 & 0.890$\pm$0.017 & 4.39$\pm$0.92 & \textbf{0.924$\pm$0.008} & \textbf{3.11$\pm$0.79} & 0.920$\pm$0.016 & 3.31$\pm$1.01 \\ \hline
Cau.L & 0.808$\pm$0.079 & 9.89$\pm$3.09 & 0.797$\pm$0.046 & 3.56$\pm$1.30 & 0.885$\pm$0.074 & 3.44$\pm$1.89 &\textbf{ 0.894$\pm$0.071} & \textbf{3.32$\pm$2.00} \\ \hline
Cau.R & 0.801$\pm$0.042 & 10.39$\pm$3.09 & 0.837$\pm$0.117 & 4.16$\pm$1.37 & 0.887$\pm$0.065 & 3.60$\pm$1.67 & \textbf{0.892$\pm$0.057} & \textbf{3.51$\pm$1.67} \\ \hline
Put.L & 0.771$\pm$0.039 & 6.31$\pm$1.09 & 0.860$\pm$0.060 & 3.79$\pm$1.76 & 0.909$\pm$0.042 & 3.07$\pm$1.40 & \textbf{0.916$\pm$0.023} & \textbf{2.63$\pm$1.09} \\ \hline
Put.R & 0.799$\pm$0.026 & 5.85$\pm$0.84 & 0.876$\pm$0.080 & 3.26$\pm$1.23 & 0.908$\pm$0.046 & 2.91$\pm$1.41 & \textbf{0.914$\pm$0.031} & \textbf{2.75$\pm$0.99} \\ \hline
Pal.L & 0.693$\pm$0.189 & 3.89$\pm$1.07 & 0.815$\pm$0.088 & 2.89$\pm$0.71 & \textbf{0.873$\pm$0.032} & 2.52$\pm$0.54 & 0.843$\pm$0.101 & \textbf{2.38$\pm$0.76} \\ \hline
Pal.R & 0.792$\pm$0.085 & 3.45$\pm$0.98 & 0.799$\pm$0.060 & 3.18$\pm$0.93 & \textbf{0.874$\pm$0.047} & \textbf{2.49$\pm$0.59} & 0.861$\pm$0.049 & 2.59$\pm$0.61 \\ \hline
Hip.L & 0.784$\pm$0.054 & 6.35$\pm$1.87 & 0.809$\pm$0.022 & 5.49$\pm$1.66 & 0.871$\pm$0.024 & \textbf{4.34$\pm$1.66} & \textbf{0.876$\pm$0.020} & 4.48$\pm$2.02 \\ \hline
Hip.R & 0.794$\pm$0.025 & 6.19$\pm$1.59 & 0.810$\pm$0.140 & 4.80$\pm$1.66 & 0.869$\pm$0.022 & 4.01$\pm$1.45 & \textbf{0.879$\pm$0.020} & \textbf{3.76$\pm$1.23} \\ \hline
Amy.L & 0.585$\pm$0.064 & 5.05$\pm$0.97 & 0.721$\pm$0.053 & 3.54$\pm$0.72 & 0.832$\pm$0.026 & \textbf{2.44$\pm$0.29} & \textbf{0.833$\pm$0.032} & 2.39$\pm$0.39 \\ \hline
Amy.R & 0.576$\pm$0.076 & 5.43$\pm$0.90 & 0.707$\pm$0.054 & 4.11$\pm$0.75 & 0.812$\pm$0.033 & \textbf{2.72$\pm$0.50} & \textbf{0.821$\pm$0.027} & 2.72$\pm$0.69 \\ \hline
Acc.L & 0.630$\pm$0.055 & 4.28$\pm$1.11 & 0.699$\pm$0.089 & 6.81$\pm$8.76 & 0.790$\pm$0.050 & 2.57$\pm$0.67 & \textbf{0.799$\pm$0.052} & \textbf{2.39$\pm$0.64} \\ \hline
Acc.R & 0.443$\pm$0.065 & 5.47$\pm$1.02 & 0.678$\pm$0.081 & 3.93$\pm$1.75 & 0.783$\pm$0.058 & 2.65$\pm$0.76 & \textbf{0.791$\pm$0.067} & \textbf{2.54$\pm$0.65} \\ \hline \hline
Avg. & 0.725$\pm$0.137 & 5.87$\pm$2.48 & 0.799$\pm$0.094 & 4.18$\pm$2.76 & 0.867$\pm$0.061 & 3.08$\pm$1.27 & \textbf{0.869$\pm$0.064} & \textbf{3.01$\pm$1.30} \\ \hline
\end{tabular}
\end{table*}
Table~\ref{table:mean_dsc_hd_miccai} shows overall and per structure mean DSC and HD values  on the MICCAI 2012 dataset. According to the results, our method showed significantly higher DSC of 0.869 $(p < 0.001)$ than FIRST and FreeSurfer which yielded 0.799 and 0.725 overall mean DSC, respectively. Moreover, as it can be observed, the HD values showed similar behavior as DSC, where the proposed approach significantly outperformed both of these methods $(p < 0.001)$, in average, with a reduction of 1.17 mm and 2.86 mm with respect to FIRST and FreeSurfer. Our method did not show a significant difference in comparison with PICSL $(p > 0.05)$ in terms of DSC having similar mean of 0.867 and 0.869 for PICSL and our method, respectively. The HD values of our approach and PICSL also confirmed previously observed DSC numbers.

\begin{figure}[tp]
\centering
\includegraphics[width=0.48\textwidth]{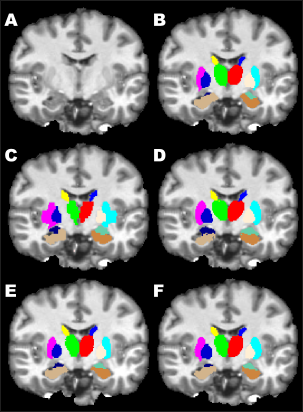}
\caption{Qualitative comparison of segmentation outputs obtained by FreeSurfer, FIRST, PICSL, and our method on MICCAI 2012 dataset. A) T1-w image; B) Groundtruth; C) FreeSurfer; D) FIRST; E) PICSL; F) Our method. Visible structures on coronal view: thalamus, caudate, pallidum, putamen, hippocampus, and amygdala.}
\label{fig:miccai_qualitative_comp}
\end{figure}

Figure~\ref{fig:miccai_qualitative_comp} shows a qualitative comparison of segmentation outputs from FreeSurfer, FIRST, PICSL, and our method. As it can be observed, FreeSurfer provided the worst segmentation output with coarse structure boundaries. FIRST produced smooth segmentation on the borders, however, the overlap between the groundtruth was poor. Our method's segmentation output was similar to the one of PICSL's and both of the methods had consistent structure boundaries, which were not far from the groundtruth. 

Apart from having similar results to the best performing method on this dataset, our strategy gained a good improvement in training and segmentation times. According to \cite{landman2012miccai}, PICSL took 330 CPU hours for training 138 classifiers used for correcting systematic errors. Reported segmentation time of PICSL with optimal parameters was more than $50$ minutes per subject volume \citep{wang2013picsl}. In comparison with the above, the execution time of our CNN strategy was around 8 hours for training and less than 5 minutes for testing, including the atlas registration.

\subsection{IBSR 18 Dataset} \label{sec:ibsr_results}
This dataset consists of 18 T1-w subject volumes with manually segmented ground truth with 32 classes. Similarly to the MICCAI 2012 dataset, we extracted 14 classes corresponding to seven sub-cortical brain structures with left and right parts separately. The subject volumes of this dataset have dimensionality of $256\times 256\times 128$ and different voxel spacings: $0.84\times 0.84\times 1.5$ mm$^3$, $0.94\times 0.94\times 1.5$ mm$^3$, and $1.00\times 1.00\times 1.5$ mm$^3$. Images in this dataset have lower contrast and resolution in comparison with the MICCAI 2012 dataset, which makes the segmentation task even more challenging.

\subsubsection{Experimental details}
For the experiments with this dataset, we followed the same preprocessing steps as done with the MICCAI 2012 dataset, which included skull-stripping and bias field correction. Since there was no training and testing split on this dataset, we performed our experiments using a leave-one-subject-out cross-validation scheme. 
For each 17-1 fold, we extracted around $1.1M$ patches from each of the three orthogonal views, divided into $825K$ ($75\%$) training and $220K$ ($25\%$) validation sets. Each model was trained for 200 epochs applying also early stopping in the training process after 20 epochs. 

\subsubsection{Comparison with other available methods}
For this dataset, the comparison of our results will be shown: 1) with the state-of-the-art FreeSurfer and FIRST methods including the statistical significance test, since the  evaluation values for each subject volume were computed by us using the corresponding tools; and 2) with recent CNN approaches of 
\cite{shakeri2016sub}, \cite{mehta2017brainsegnet} (BrainSegNet), \cite{bao2016multi} (MS-CNN), and \cite{dolz20173d}. The results for the recent methods were taken from their corresponding papers exactly as they have been reported. We have to mention that most of the CNN based methods report results only for a specific group of sub-cortical structures, but do not show or consider the results for the other, yet important, sub-cortical structures. 
Note also that the comparison on HD metric is present only for FreeSurfer, FIRST and our method, but not for other considered methods because most of the approaches do not report HD values.

\subsubsection{Results}
Table~\ref{table:ibsr_methods_comparison} shows the mean DSC and HD values for each of the evaluated methods. Our method showed a better performance in comparison to both FreeSurfer and FIRST methods for all the sub-cortical structures. The overall DSC mean of our method was significantly higher than both of the methods $(p < 0.001)$, with mean DSC of 0.740, 0.808, and 0.843 for FreeSurfer, FIRST and the proposed strategy, respectively. In terms of HD values, our method showed overall mean of $4.49$, whereas FreeSurfer and FIRST yielded $5.21$ and $4.50$, respectively. The proposed strategy significantly outperformed FreeSurfer with $(p < 0.001)$, however the difference with FIRST was not significant $(p > 0.5)$. As shown in Table~\ref{table:ibsr_methods_comparison}, FreeSurfer performed worst for almost all the structures, while FIRST and our method showed similar performance. On both thalamus structures, our method showed lowest score in comparison with the other methods, however it yielded better HD for the small structures like amygdala, accumbens, and hippocampus. In general, HD metric is very sensitive to outliers, hence, a few misclassified voxels can cause considerable reduction in performance as seen in the results for the thalamus structure in our method.

\begin{table*}[tb]
\scriptsize
\centering
\caption{Comparison of our method with the state-of-the-art methods as well as previous CNN approaches on IBSR dataset in terms of DSC, HD, and standard deviation. Structure acronyms are: left thalamus (Tha.L), right thalamus (Tha.R), left caudate (Cau.L), right caudate (Cau.R), left putamen (Put.L), right putamen (Put.R), left pallidum (Pal.L), right pallidum (Pal.R), left hippocampus (Hip.L), right hippocampus (Hip.R), left amygdala (Amy.L), right amygdala (Amy.R), left accumbens (Acc.L), right accumbens (Acc.R). ``-'' represents no results were reported on corresponding structure. The average (Avg.) values show mean DSC for the presented structure DSC scores. Highest DSC and HD values for each structure are shown in bold.}
\label{table:ibsr_methods_comparison}
\setlength\tabcolsep{5pt}
\begin{tabular}{|c|c|c|c|c|c|c|c|c|c|c|}
\hline
Method & \multicolumn{2}{c|}{FreeSurfer} & \multicolumn{2}{c|}{FIRST} & \cite{shakeri2016sub} & BrainSegNet & MS-CNN & \cite{dolz20173d} & \multicolumn{2}{c|}{Our method} \\ \hline
Struct. & DSC & HD & DSC & HD & DSC & DSC & DSC & DSC & DSC & HD \\ \hline \hline
Tha.L & 0.815$\pm$0.056 & 5.367$\pm$1.168 & 0.893$\pm$0.017 & \textbf{3.819$\pm$0.850} & 0.866$\pm$0.023 & 0.88$\pm$0.050 & \multirow{2}{*}{0.889} & \multirow{2}{*}{\textbf{0.92}} & 0.910$\pm$0.014 & 7.159$\pm$0.402 \\ \cline{1-7} \cline{10-11} 
Tha.R & 0.864$\pm$0.022 & \textbf{4.471$\pm$1.245} & 0.885$\pm$0.012 & 4.273$\pm$1.137 & 0.874$\pm$0.021 & 0.90$\pm$0.029 &  &  & 0.914$\pm$0.016 & 7.256$\pm$0.571 \\ \hline
Cau.L & 0.796$\pm$0.050 & 6.435$\pm$1.939 & 0.783$\pm$0.044 & 4.128$\pm$1.575 & 0.778$\pm$0.053 & 0.86$\pm$0.047 & \multirow{2}{*}{0.849} & \multirow{2}{*}{\textbf{0.91}} & 0.896$\pm$0.018 & \textbf{4.054$\pm$1.412} \\ \cline{1-7} \cline{10-11} 
Cau.R & 0.809$\pm$0.048 & 8.201$\pm$2.443 & 0.870$\pm$0.027 & \textbf{3.687$\pm$0.791} & 0.783$\pm$0.068 & 0.88$\pm$0.048 &  &  & 0.896$\pm$0.020 & 4.153$\pm$1.061 \\ \hline
Put.L & 0.789$\pm$0.038 & 5.310$\pm$0.923 & 0.869$\pm$0.020 & \textbf{4.421$\pm$1.185} & 0.838$\pm$0.026 & \textbf{0.91$\pm$0.022} & \multirow{2}{*}{0.875} & \multirow{2}{*}{0.90} & 0.900$\pm$0.014 & 5.216$\pm$1.788 \\ \cline{1-7} \cline{10-11} 
Put.R & 0.829$\pm$0.031 & 4.716$\pm$1.189 & 0.880$\pm$0.010 & 4.725$\pm$1.814 & 0.824$\pm$0.039 & \textbf{0.91$\pm$0.023} &  &  & 0.904$\pm$0.012 & \textbf{4.577$\pm$0.410} \\ \hline
Pal.L & 0.632$\pm$0.171 & 4.652$\pm$1.294 & 0.810$\pm$0.033 & \textbf{3.477$\pm$0.572} & 0.763$\pm$0.031 & 0.81$\pm$0.089 & \multirow{2}{*}{0.787} & \multirow{2}{*}{\textbf{0.86}} & 0.825$\pm$0.050 & 3.849$\pm$0.574 \\ \cline{1-7} \cline{10-11} 
Pal.R & 0.774$\pm$0.032 & 3.966$\pm$0.793 & 0.809$\pm$0.037 & 3.990$\pm$1.075 & 0.736$\pm$0.055 & 0.83$\pm$0.086 &  &  & 0.829$\pm$0.046 & \textbf{3.700$\pm$0.576} \\ \hline
Hip.L & 0.760$\pm$0.036 & 5.787$\pm$1.264 & 0.806$\pm$0.023 & 5.571$\pm$1.592 & - & 0.81$\pm$0.065 & \multirow{2}{*}{0.788} & \multirow{2}{*}{-} & \textbf{0.851$\pm$0.024} & \textbf{4.177$\pm$1.087} \\ \cline{1-7} \cline{10-11} 
Hip.R & 0.767$\pm$0.060 & 5.615$\pm$1.600 & 0.817$\pm$0.023 & 4.349$\pm$0.984 & - & 0.83$\pm$0.071 &  &  & \textbf{0.851$\pm$0.024} & \textbf{4.124$\pm$0.824} \\ \hline
Amy.L & 0.661$\pm$0.069 & 5.521$\pm$1.517 & 0.742$\pm$0.064 & 4.648$\pm$1.950 & - & 0.76$\pm$0.087 & \multirow{2}{*}{0.654} & \multirow{2}{*}{-} & \textbf{0.763$\pm$0.052} & \textbf{4.326$\pm$0.822} \\ \cline{1-7} \cline{10-11} 
Amy.R & 0.690$\pm$0.067 & 4.720$\pm$1.553 & 0.757$\pm$0.062 & 4.402$\pm$1.493 & - & 0.71$\pm$0.087 &  &  & \textbf{0.768$\pm$0.058} & \textbf{4.292$\pm$1.064} \\ \hline
Acc.L & 0.604$\pm$0.071 & 3.634$\pm$0.783 & 0.684$\pm$0.098 & 7.770$\pm$8.803 & - & - & \multirow{2}{*}{-} & \multirow{2}{*}{-} & \textbf{0.744$\pm$0.053} & \textbf{3.026$\pm$0.676} \\ \cline{1-7} \cline{10-11} 
Acc.R & 0.574$\pm$0.074 & 4.507$\pm$1.077 & 0.703$\pm$0.076 & 3.733$\pm$1.482 & - & - &  &  & \textbf{0.752$\pm$0.047} & \textbf{2.995$\pm$0.609} \\ \hline \hline
Avg. & 0.740$\pm$0.110 & 5.207$\pm$1.761 & 0.808$\pm$0.080 & 4.499$\pm$2.810 & 0.808$\pm$0.063 & 0.841$\pm$0.064 & 0.807 & 0.898 & 0.843$\pm$0.071 & 4.493$\pm$1.533 \\ \hline
\end{tabular}
\end{table*}

Compared to other CNNs, our approach outperformed the method proposed by Shakeri \textit{et~al.} (DSC = $0.808$) on the eight evaluated structures. Similarly, the performance of the proposed approach was also superior on the six structures evaluated in the work of Mehta \textit{et~al.} (DSC=$0.841$). Further, we compare our method with MS-CNN, which has reported average DSC values for six structures for left and right parts together (overall DSC = $0.807$). Our method's mean DSC on these structures was $0.859$, which was higher than the result of MS-CNN ($0.807$) and yielded higher DSC scores for all the structures. Finally, when compared with the work of Dolz \textit{et~al.}, our method showed a comparable performance, although this last work showed slightly higher averaged DSC values for the four biggest structures.

\subsection{Effect of the spatial priors} \label{sec:effect_of_the_spatial_priors}
We ran experiments using the proposed method with and without spatial priors to determine the effect of using such features to the segmentation performance on both datasets. For this experiment, we analyzed the results in terms of DSC on the MICCAI 2012 dataset. We did not present the results of this experiment for the IBSR 18 dataset for simplicity, since it produced the similar outcome. In order to test our network without the spatial features, we modified the architecture (Figure~\ref{fig:architecture_proposal}) by removing the branch of atlas probabilities and keeping only three branches of convolutional layers.

\begin{figure}[tb]
\centering
\includegraphics[width=0.48\textwidth]{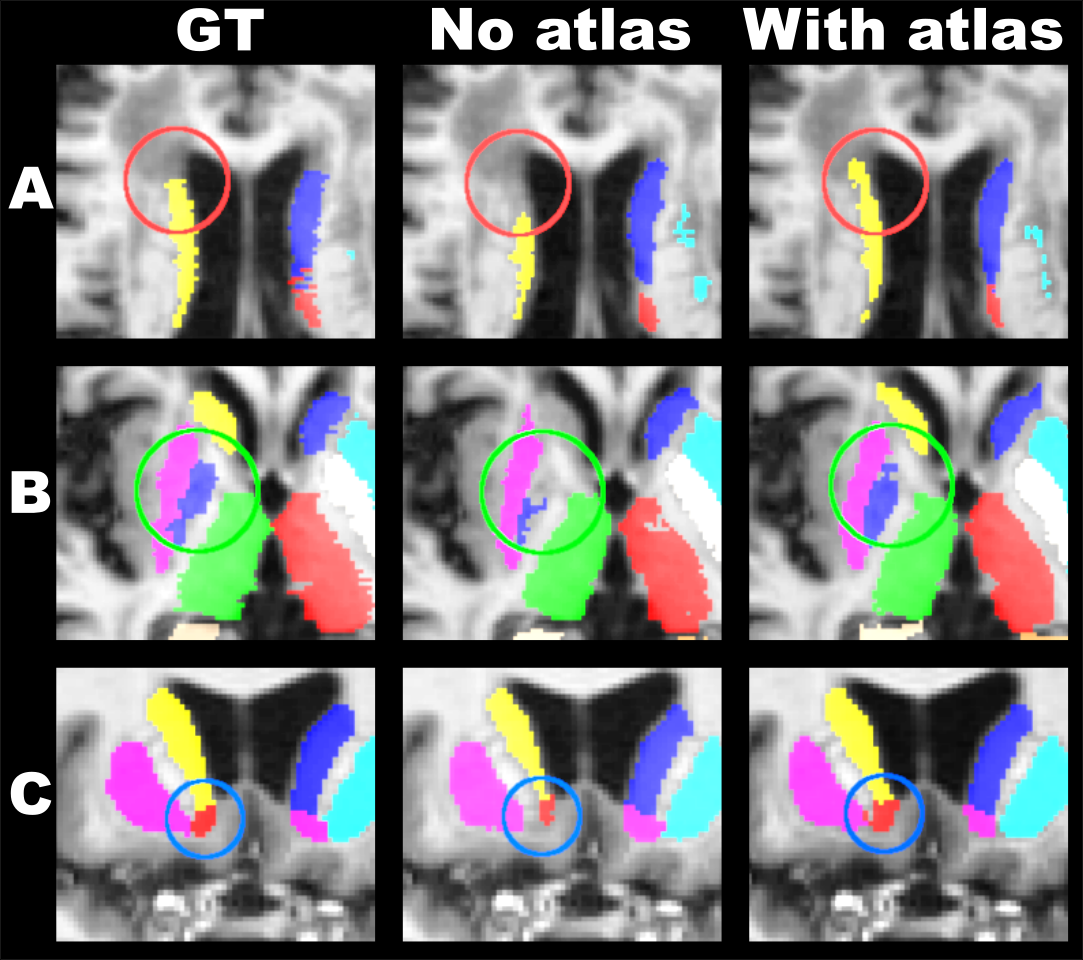}
\caption{Comparison of segmentation output for the difficult areas of the (A) caudate, (B) pallidum, and (C) accumbens structures in some of the images from MICCAI 2012 dataset using the proposed method with and without using the spatial priors. The caudate and pallidum areas are shown in red and green circles respectively from axial view, and accumbens is shown in blue circle from coronal view.}
\label{fig:difficult_image_segmentation_comparison}
\end{figure}

Table~\ref{table:effect_of_atlas_and_sampling},
\begin{table}[tb]
\scriptsize
\centering
\caption{Effect of spatial features and the proposed sample selection technique. MICCAI 2012 dataset. Random sampling -- method without using the sample selection from boundaries (including the spatial priors). No atlas -- method without incorporating atlas priors (using the sampling technique). Final method -- proposed method that includes both the spatial features and the sampling technique. Structure acronyms are: left thalamus (Tha.L), right thalamus (Tha.R), left caudate (Cau.L), right caudate (Cau.R), left putamen (Put.L), right putamen (Put.R), left pallidum (Pal.L), right pallidum (Pal.R), left hippocampus (Hip.L), right hippocampus (Hip.R), left amygdala (Amy.L), right amygdala (Amy.R), left accumbens (Acc.L), right accumbens (Acc.R).}
\label{table:effect_of_atlas_and_sampling}
\begin{tabular}{|c|c|c|c|}
\hline
Method & Random sampling & No atlas & Final method \\ \hline \hline
Tha.L & 0.860$\pm$0.013 & 0.911$\pm$0.024 & \textbf{0.921$\pm$0.017} \\ \hline
Tha.R & 0.862$\pm$0.014 & 0.917$\pm$0.017 & \textbf{0.920$\pm$0.016} \\ \hline
Cau.L & 0.831$\pm$0.067 & 0.880$\pm$0.103 & \textbf{0.894$\pm$0.071} \\ \hline
Cau.R & 0.834$\pm$0.048 & 0.864$\pm$0.131 &\textbf{ 0.892$\pm$0.057} \\ \hline
Put.L & 0.871$\pm$0.024 & 0.900$\pm$0.073 & \textbf{0.916$\pm$0.023} \\ \hline
Put.R & 0.872$\pm$0.027 & 0.913$\pm$0.029 & \textbf{0.914$\pm$0.031} \\ \hline
Pal.L & 0.784$\pm$0.040 & \textbf{0.852$\pm$0.086} & 0.843$\pm$0.101 \\ \hline
Pal.R & 0.775$\pm$0.057 & 0.833$\pm$0.099 & \textbf{0.861$\pm$0.049} \\ \hline
Hip.L & 0.778$\pm$0.034 & 0.871$\pm$0.019 & \textbf{0.876$\pm$0.020} \\ \hline
Hip.R & 0.770$\pm$0.026 & 0.876$\pm$0.018 & \textbf{0.879$\pm$0.020} \\ \hline
Amy.L & 0.709$\pm$0.025 & 0.824$\pm$0.037 & \textbf{0.833$\pm$0.032} \\ \hline
Amy.R & 0.716$\pm$0.054 & 0.819$\pm$0.035 & \textbf{0.821$\pm$0.027} \\ \hline
Acc.L & 0.744$\pm$0.060 & 0.796$\pm$0.052 & \textbf{0.799$\pm$0.052} \\ \hline
Acc.R & 0.689$\pm$0.091 & 0.753$\pm$0.106 & \textbf{0.791$\pm$0.067} \\ \hline \hline
Avg. & 0.792$\pm$0.076 & 0.858$\pm$0.083 & \textbf{0.869$\pm$0.064} \\ \hline
\end{tabular}
\end{table}
shows DSC results of our method with random sampling, without using spatial features, and the final method. Inclusion of the spatial features significantly improved the overall DSC $(p < 0.001)$, as well as the results for almost all the structures. The segmentation difference can be seen from
Figure~\ref{fig:difficult_image_segmentation_comparison},
where difficult areas of the caudate, pallidum, and accumbens structures were segmented better by the method that comprised the spatial features. Hence, the spatial priors helped to overcome difficult areas, producing more accurate segmentation for some images that had intensity and shape irregularities that could not be observed in any of the training images.

\subsection{Effect of sample selection}
In this section, we show the effect of sample selection from structure boundaries using the MICCAI 2012 dataset. For this experiment, random sample selection from all the brain tissues has been used for training the network. For every epoch, we extracted the same number of voxels ($1.5M$) for both the sub-cortical structures ({$750K$}) and background ({$750K$}). Here, background voxels were randomly selected from whole brain volume, instead of selecting only from structure boundaries (see Figure \ref{fig:sample_selection_gt}).  The network was again trained for 200 epochs using the same configuration. Spatial features were also included in training.

\begin{figure}[tb]
\centering
\includegraphics[width=0.45\textwidth]{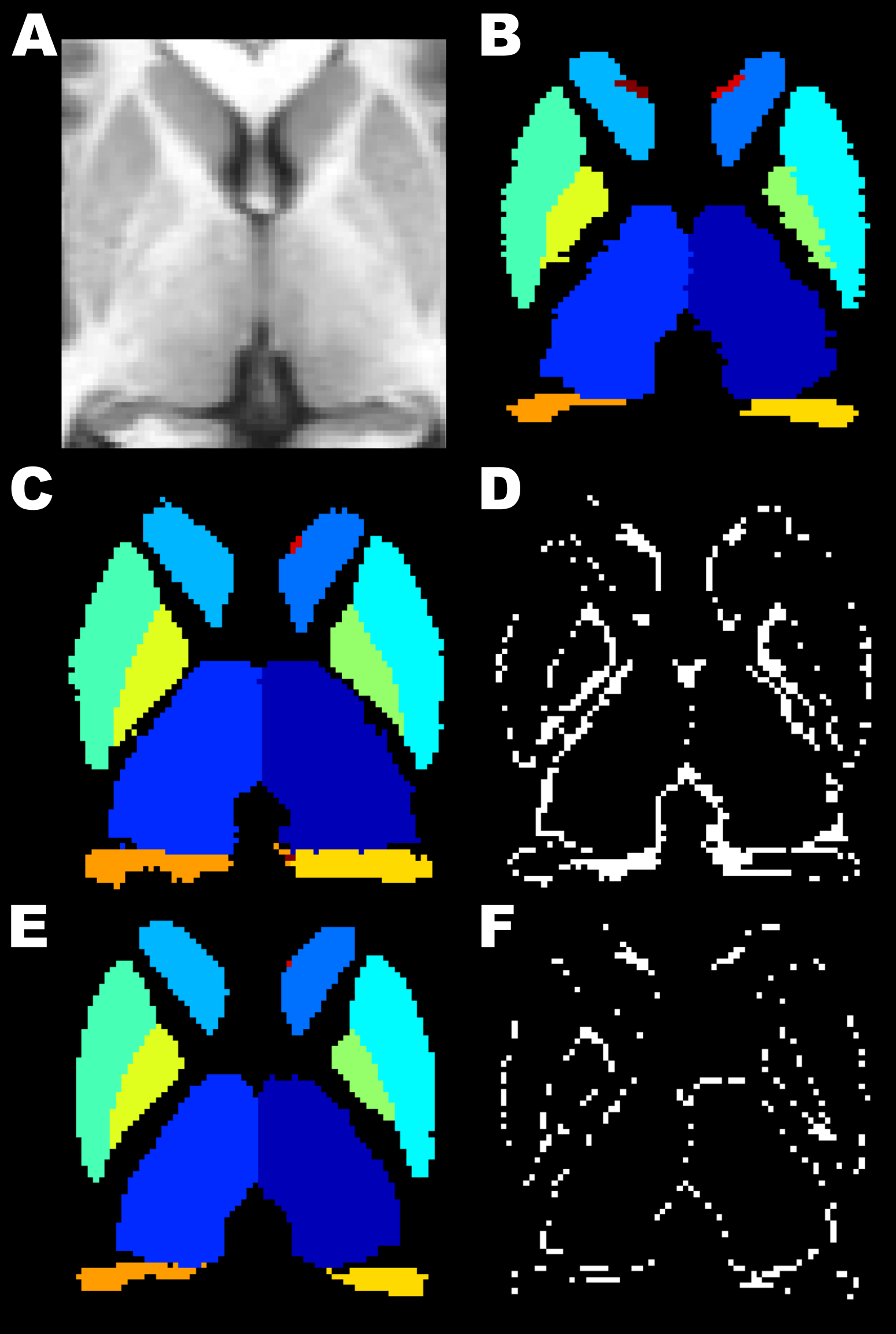}
\caption{Illustration of misclassification occurrence on borders. MICCAI 2012 dataset. (A, B) T1-w image and manual segmentation; (C, D) segmentation using random sample selection and difference from groundtruth; (E, F) segmentation using the sample selection from borders and difference from groundtruth.}
\label{fig:border_difference}
\end{figure}

Table~\ref{table:effect_of_atlas_and_sampling} shows the results corresponding to this experiment. Mean DSC obtained with our network without using the sample selection technique was $0.792$ compared to $0.869$ of the final approach. Accordingly, the proposed sample selection technique significantly improved the network's performance in average as well as for each of the structures $(p < 0.001)$ . Figure~\ref{fig:border_difference} illustrates the segmentation results produced by our final approach and without applying sampling from borders. As it can be seen from the difference between groundtruth and segmentation masks, the final strategy produced better segmentation on the boundaries than random sample selection method.
In fact, the difference of our segmentation and the ground truth mask was not substantial, but only a few voxels. We also can observe that the intensities on the border voxels of the structures are mostly confounding. Therefore, assigning these voxels to the structure or background is highly dependent on ground truth.

\section{Discussion} \label{sec:discussion}
In this paper, we have proposed a fully automated 2.5D patch-based CNN approach that combines both convolutional and a priori spatial features for accurate segmentation of the sub-cortical brain structures. In our approach, a structural sub-cortical atlas has been registered into the image space to extract the spatial probability of each voxel. Then, fused with the extracted convolutional features in the fully connected layers. The inclusion of the spatial information increases the execution time by adding atlas registration. However, it allows us to filter out misclassified regions that have bigger size than the actual structures in the segmentation output, which may appear in unobserved areas (i.e. not included in the training phase) of the brain as a consequence of applying restricted sampling.
As seen in all the experiments, the addition of the spatial priors and the restricted sampling strategy have a significant effect on the accuracy of the proposed method, outperforming or showing a comparable performance to both classic as well as other novel learning approaches for segmenting the sub-cortical structures. 

Compared to other state-of-the-art techniques such as FreeSurfer and FIRST, the spatial agreement of the proposed method with the manual segmentation is clearly higher in all evaluated datasets. As seen in other radiological tasks, this reinforces the effectiveness of CNN techniques when manual expert annotations are available. On the MICCAI 2012 dataset, our method shows an excellent performance, slightly over-performing the best challenge participant strategy -- PICSL. Although not directly evaluated, our method clearly reduces the training and inference time. However, it has to be noted that most of the execution time of PICSL is due to highly computational registration processes which were carried out on CPU, while our method relies on GPU processors to speed-up training. Other CNN methods have also been evaluated on the MICCAI 2012 database \citep{wachinger2017deepnat,mehta2017brainsegnet}. However, these works do not report exact evaluation values for sub-cortical structures, constraining us in performing a quantitative comparison.

In contrast, different CNN methods that have been evaluated using the IBSR 18 dataset have reported exact numerical values. When compared to other CNN approaches, our method also showed a significant increase in the performance with respect to most of them, and a comparable performance with the method proposed by Dolz \textit{et~al}. However, as seen in Section~\ref{sec:ibsr_results}, previous studies do not always deal with all sub-cortical structures, restricting a more detailed comparison with respect to other methods. Additionally, the training methodology also differed among the strategies. In this aspect, although all our experiments were carried out using the leave-one-out approach, we also repeated our IBSR 18 experiments using a 6-fold (15 training and 3 testing) validation strategy to perform a fair comparison with some of the considered methods. The complete results of the 6-fold validation strategy were not depicted in the paper for simplicity, but, our network achieved similar results with only $0.005$ of difference in DSC with respect to the leave-one-out strategy, showing the robustness of the proposed approach to changes in the number of training images.

According to the experimental results, employing the spatial features to the CNN significantly improved the performance of the network. The atlas priors showed to be useful in guiding the network when segmenting the difficult areas. As we have seen in Section~\ref{sec:effect_of_the_spatial_priors}, CNN that leveraged the spatial priors coped with these intensity based difficulties. Accordingly, by providing the atlas probabilities, we make sure that the anatomical shape and structure are taken into account before assigning a label to a voxel. Since the sub-cortical structures follow the similar anatomical structure in all patients, the inclusion of the spatial features makes the segmentation approach more robust to irregularities in intensity based features obtained from T1-w images by providing additional location-based information. 
Despite being prone to the inherent errors in image registration, the addition of these a priori spatial class probabilities, or other explicit fused problem-specific information, may have other direct benefits such as reduction of the effect of low contrast, poor resolution, presence of noise, and artifacts close to the structure boundaries.

Our results also show the importance of sampling and class balancing in the training process. By feeding the network with only the most difficult negative samples, we ensure that useful samples were used in the training process. When compared to the rest of CNN approaches, our method without restricted sampling yields a similar performance to other methods such as the one of \cite{shakeri2016sub} and MS-CNN \citep{bao2016multi} even if trained on the same conditions, which highlights the effectiveness of the used sampling strategy. As a counterpart, this kind of approaches tend to generate false positive regions outside the sub-cortical space, due to the lack of contextual spatial information of the whole brain. Within our approach, we take advantage of the already computed spatial priors to reduce the segmentation to only a region of interest containing the sub-cortical structures, which reduces remarkably the inference time. Remaining false positive voxels are then post-processed by maintaining only the biggest region for each class.


Our study comprises some limitations. As part of supervised training strategies, the accuracy of CNN methods tend to decrease significantly in other image domains (i.e. different MRI scanner, image protocol, etc.) than the ones used for training. Nevertheless, there is still a little evidence of the capability of CNN methods in radiological tasks with small or none datasets, which highlights the need of further studying this issue to increase the accuracy of such approaches. With no more evidence in this field, FIRST may be more appropriate in these scenarios when few or no training data is available. Another constraint involves the applicability of the proposed method on datasets of images with neurological diseases comprising, for instance, white matter lesions, which has been recently shown in \cite{gonzalez2017evaluating}, where such conditions affected the brain structure segmentation task.


\section{Conclusion} \label{sec:conclusion}
In this paper, we have presented a novel CNN based deep learning approach for accurate and robust segmentation of the sub-cortical brain structures that combines both convolutional and prior spatial features for improving the segmentation accuracy. In order to increase the accuracy of the classifier, we have proposed to train the network using a restricted sample selection to force the network to learn the most difficult parts of the structures.
As seen from all the experiments carried out on the public MICCAI 2012 and IBSR 18 datasets, the addition of the spatial priors and the restricted sampling strategy have a significant impact on the effectiveness of the proposed method, outperforming or showing a comparable performance to state-of-the-art methods such as FreeSurfer, FIRST and different recently proposed CNN approaches. In order to encourage the reproducibility and the use of the proposed method, a public version is available to download for the neuroimaging community at our research website.

\section{Acknowledgements}
Kaisar Kushibar and Jose Bernal hold FI-DGR2017 grant from the Catalan Government with reference numbers 2017FI\_B00372 and 2017FI\_B00476 respectively. This work has been partially supported by La Fundaci\'o la Marat\'o de TV3, by Retos de Investigación TIN2014-55710-R and TIN2015-73563-JIN from the Ministerio de Ciencia y Tecnologia, and by MPC UdG 2016/022 grant. The authors gratefully acknowledge the support of the NVIDIA Corporation with their donation of the TITAN-X PASCAL GPU used in this research.

\section*{References}

\end{document}